\documentclass{article}
\usepackage{spconf,amsmath,graphicx}
\usepackage{amsfonts, amssymb, bbding, pifont, utfsym}
\usepackage{booktabs,multirow}
\usepackage{color}
\usepackage{makecell}

\DeclareMathOperator*{\argmin}{arg\,min}

\title{
Extending Whisper with Prompt Tuning to Target-Speaker ASR}
\name{Hao Ma$^{1}$, Zhiyuan Peng$^{2}$, Mingjie Shao$^{1}$, Jing Li$^{3}$ and Ju Liu$^{1}$ \thanks{This work was supported in part by the National Natural Science Foundation of China under Grant 62071275; in part by the Innovation and Development Joint Funds of Shandong Natural Science Foundation under Grant ZR2022LZH012 and in part by the Key Research and Development Program of China under Grant 2022YFC3302800. Corresponding authors: Ju Liu (juliu@sdu.edu.cn) and Mingjie Shao (mingjieshao@sdu.edu.cn). Our source code is available at https://github.com/Aisaka0v0/TS-Whisper.}
}
\address{\small$^{1}$School of Information Science and Engineering, Shandong University, Qingdao, China\\
\small$^{2}$Department of Computer Science, North Carolina State University, North Carolina, USA\\
\small$^{3}$School of Journalism and Communication, Shandong Normal University, Jinan, China}
\begin{document}
\ninept
\maketitle
\begin{abstract}
Target-speaker automatic speech recognition (ASR) aims to transcribe the desired speech of a target speaker from multi-talker overlapped utterances. 
Most of the existing target-speaker ASR  (TS-ASR) methods involve either training from scratch or fully fine-tuning a pre-trained model, leading to significant training costs and becoming inapplicable to large foundation models. 
This work leverages prompt tuning, a parameter-efficient fine-tuning approach, to extend Whisper, a large-scale single-talker ASR model, to TS-ASR. 
Variants of prompt tuning approaches along with their configurations are explored and optimized for TS-ASR.
Experimental results show that prompt tuning can achieve performance comparable to state-of-the-art full training approaches while only requiring about 1\% of task-specific model parameters. 
Notably, the original Whisper's features, such as inverse text normalization and timestamp tagging, are retained in target-speaker ASR, keeping the generated transcriptions natural and informative.
\end{abstract}
\section{Introduction}
\label{sec:intro}
\begin{figure*}
  \centering
  \includegraphics[width=15cm]{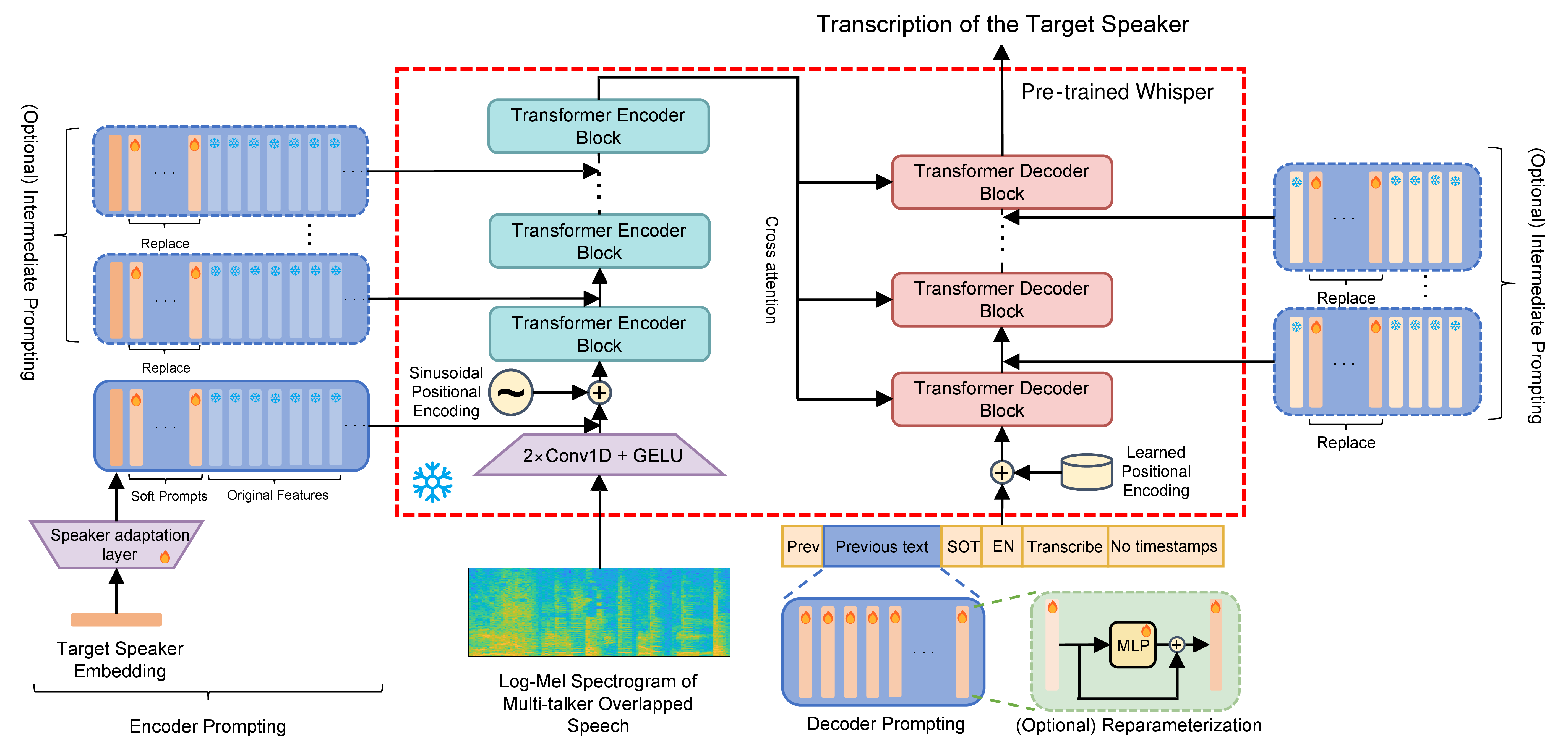}
  \vspace{-5mm}
  \caption{\it Overview of proposed prompting framework.
Modules with solid-line borders are included in the baseline configuration, while modules with dashed-line borders are optional.}
  \label{fig1}
  \vspace{-7mm}
\end{figure*}
With the development of deep learning techniques, existing automatic speech recognition (ASR) models have already reached or even surpassed the human level under single-speaker conditions \cite{whisper}. 
However, it remains challenging to extend ASR models to transcribe overlapped speech from multi-talkers, which commonly exist in many practical scenes. The interference between speakers' voices makes a single-talker ASR model confused about which speaker's speech to transcribe. 
To handle this problem, multi-talker ASR systems become the focus of much attention. 

Several representative methods have been developed with the progress of deep learning technologies.
Initial multi-talker ASR systems are built based on speech separation, where each individual speaker's voice is separated first, followed by single-speaker ASR for transcription. 
Such a cascading system, though can be comparatively feasible to build for each individual component, achieves very limited performance. This is because speech separation is not directly optimized for ASR. 
In this regard, end-to-end multi-talker ASR  is preferred. 
Without the need for explicitly separating each speaker's speech, the overall computation cost can be greatly reduced.
Following this direction, techniques such as permutation invariant training (PIT) \cite{PIT, qian_single-channel_2018, sidecar} and serialized output training (SOT) \cite{sot, tuning_whisper} are developed, allowing the joint transcription for all speakers in overlapped utterances without any auxiliary speaker information. 
Meanwhile, speaker-attribute ASR (SA-ASR) \cite{saasr} leverages speaker-dependent clues to augment the transcription of overlapped utterances for improved performance. Another research direction related to SA-ASR focuses on transcribing only the desired speech of a specific target speaker, known as \textit{target-speaker ASR} (TS-ASR) \cite{tse2_baseline, tse3, tse4_baseline, tse5_baseline}. In the context of TS-ASR, only the target speaker's profile is required, which can be convenient to acquire.

However, many of the existing TS-ASR methods either train a specially designed model from scratch \cite{tse2_baseline, tse3} or fully fine-tune a pre-trained model \cite{tse4_baseline, tse5_baseline}. 
Though achieving SOTA performance, training from scratch and fully fine-tuning require a significant amount of computation and storage. 
In the era of large foundation models, a single {\it universal model} is expected to handle multiple tasks, and when needed, can be efficiently adapted to new tasks, at small incremental computation costs.
This can be achieved by parameter-efficient fine-tuning (PEFT) techniques \cite{adapter, hu2022lora, prompt, prefix}. 
In particular, {\it prompt tuning} \cite{prompt, prefix} has attracted considerable attention in both natural language processing (NLP)~\cite{prompt, razdaibiedina2023residual} and speech processing~\cite{tuning_whisper, chang2023speechprompt, avprompt} due to its remarkable parameter efficiency and promising performance. Intrigued by its potential, in this work, we explore the application of prompt tuning in the target-speaker ASR task. To the best of our knowledge, this is the first work that studies the application of prompt tuning for target-speaker ASR.

Regarding the choice of the foundation speech model, we adopt {\it Whisper} \cite{whisper}, which has recently gained considerable attention due to its impressive performance in many speech tasks.
The remarkable transferability of Whisper \cite{tuning_whisper, promptwhisper} inspires us to extend it to TS-ASR.

To be more specific, we propose a prompt tuning approach that uses target speaker embeddings together with trainable soft prompts to guide Whisper to perform TS-ASR. 
Deep prompting \cite{p-tuning} and reparameterization \cite{razdaibiedina2023residual} techniques are also leveraged to further improve the performance. 
Along with extending Whisper to TS-ASR, our method also keeps the original Whisper's features, such as inverse text normalization and timestamp prediction.   
As a result, the generated transcriptions are more natural and informative.
Extensive experiments are carried out, showing that our proposed method is competitive with state-of-the-art (SOTA) full training and fine-tuning methods~\cite{tse4_baseline, tse5_baseline}, while 
tuning and storing only about 1\% of the task-specific model parameters. 
\vspace{-2mm}
\section{Methods}
\label{sec:Methods}
We adopt prompt tuning to extend Whisper to target-speaker ASR (TS-ASR). 
An overview of the proposed framework is shown in Fig.\ref{fig1}, which is composed of four components: 1) a pre-trained foundation model---Whisper, to perform speech recognition, 2) encoder and decoder prompting, which precedes a target speaker embedding and trainable soft prompts to the input features (or text tokens) for target-speaker ASR, 3) intermediate prompting, which replaces the preceding output vectors from intermediate blocks with soft prompts for performance improvement, 4) reparameterization, which reparameterizes all the preceding soft prompts with residual neural networks for training stability.
\vspace{-3mm}
\subsection{Background of Whisper}
\label{sec2.1}
Whisper \cite{whisper} is a powerful encoder-decoder Transformer that is capable of multiple speech tasks, including multilingual speech recognition, phrase-level timestamp prediction, inverse text normalization, etc.
The input of Whisper is an 80-dimension log-Mel spectrogram $\mathbf{X} \in \mathbb{R}^{80 \times T}$ where $T$ denotes the context length. 
The spectrogram $\mathbf{X}$ is then fed into two 1-d convolution layers, where the dimension is raised to the model dimension $d_m$ and the context length is downsampled by half. 
Then the encoder blocks encode the input speech feature into hidden vectors $\mathbf{H} \in \mathbb{R}^{d_{m} \times T/2}$ and the decoder blocks decode the hidden vectors into text tokens $\hat{\mathbf{y}} $ recursively, conditioned on special tokens $\bf g$ that instruct the model for a specific speech task. Formally, this process can be described as: \\
\begin{equation*}
\begin{split}
\label{eq1}
    {\bf H} =&~ {\rm AudioEncoder}_{\phi_e}(Conv({\bf X})),\\
    \hat{y}_t = &~{\rm TextDecoder}_{\phi_d}(\mathbf{g}, \hat{y}_{1:t-1}, \mathbf{H}).
    \end{split}
\end{equation*}

For single-talker English ASR without producing timestamps, $\mathbf{g}$ is of the form: [\textit{$\langle|$prev$|\rangle$, prev-text, $\langle|$start-of-transcribe$|\rangle$, $\langle|$EN$|\rangle$, $\langle|$transcribe$|\rangle$, $\langle|$no-timestamps$|\rangle$}], where $\langle|.|\rangle$ denotes special token and \textit{prev-text} is an optional sequence of text tokens for decoding control\footnote{Refer to \cite{whisper} for more details about the multitask format of Whisper.}. 
The audio encoder and text decoder are parameterized by $\phi_e$ and $\phi_d$, respectively.

\vspace{-3mm}
\subsection{Extending Whisper to Target-speaker ASR}
Transcribing only a target speaker's speech from overlapped multi-talker utterances is challenging and beyond the capability of Whisper. 
A possible solution is prompt tuning, a representative parameter-efficient fine-tuning paradigm for foundation models. 

\subsubsection{Prompt Tuning}
Prompt tuning was first proposed in NLP to augment a frozen large language model (LLM) for new tasks \cite{gpt3, autoprompt:emnlp20}. This is achieved by introducing a limited number of leading special tokens to the input text. The special tokens function similarly to $\bf g$ in whisper, which offers the flexibility of turning a foundation model into different operation modes between multiple tasks. Moreover, these tokens are assigned with trainable token embeddings, termed \textit{soft prompts} \cite{prompt, prefix}, which are randomly initialized and optimized by stochastic gradient descent.

Formally, given the log-Mel spectrogram $\mathbf{\widetilde{X}}$ of an overlapped speech from $U$ speakers, we extend whisper to transcribe the text token sequence $\mathbf{\hat{y}}_i$ of the $i$-th speaker, conditioned on the corresponding speaker embedding $e_i\in \mathbb{R}^{d_e}$:
\begin{equation}\label{eq:AE}
\begin{split}
    \mathbf{H}_i =&~ {\rm AudioEncoder}_{\phi_e}({[\mathbf{W}e_i,\mathbf{P}_e, Conv(\mathbf{\widetilde{X}})]}),\\
    \hat{y}_{i,t} = &~{\rm TextDecoder}_{\phi_d}(\mathbf{g}(\mathbf{P}_d), \hat{y}_{i,1:t-1}, \mathbf{H}_i),
    \end{split}
\end{equation}
where $\mathbf{W} \in \mathbb{R}^{d_m \times d_e}$ is a trainable weight matrix; 
$\mathbf{P}_e \in \mathbb{R}^{d_{m} \times L_e} $ and $\mathbf{P}_d \in \mathbb{R}^{d_{m} \times L_d}$ are learnable soft prompts;
$L_e$ and $L_d$ are hyper-parameters denoting the number of soft prompts;
and $\mathbf{g}(\mathbf{P}_d)$ represents a modification of the special tokens $\mathbf{g}$, i.e., soft prompts  $\mathbf{P}_d$ are inserted into the token embeddings of the \textit{prev-text} in $\mathbf{g}$.

In the training phase, we minimize the cross-entropy loss $\mathcal{L}_{CE}$ between the model predictions $\mathbf{\hat{y}}_i$ and ground-truth transcriptions $\mathbf{y}_i$. We optimize only the speaker projection matrix $\mathbf{W}$ and soft prompts $\mathbf{P} = \{ \mathbf{P}_e, \mathbf{P}_d\}$ while keeping all the other parameters fixed:
\begin{equation}
    (\mathbf{W^*}, \mathbf{P^*}) = \argmin_{\mathbf{W},\mathbf{P}} \quad \mathcal{L}_{\text CE}(\mathbf{\hat{y}}_i,\mathbf{y}_i).
\end{equation}
\vspace{-3mm}
\subsubsection{Deep Prompting}

In Eq. \eqref{eq:AE}, soft prompts are inserted into the encoder and decoder input. 
We observe that such a simple strategy may have the following limitations:
First, it constrains the number of tunable parameters because the prompt length cannot be too long for computational cost concerns. 
Second, it is difficult to directly control the model prediction since these soft prompts are ``far'' from the output layer. 
Therefore, we further explore using deep prompting \cite{p-tuning}, wherein we replace the preceding output vectors from intermediate blocks with new soft ones for performance improvement; see Fig.~\ref{fig1} for an illustration.

\subsubsection{Reparameterization of Soft Prompts}

It has been reported that directly optimizing soft prompts may make the training process unstable, resulting in poor performance \cite{prefix, razdaibiedina2023residual}. 
To improve the stability of the training process, the reparameterization approach can be adopted~\cite{razdaibiedina2023residual}. 
It uses a residual feed-forward network to reparameterize $\mathbf{P}$ as:
\begin{equation}
\label{eq6}
    \mathbf{P'} = {\rm MLP}_{\phi_m}(\mathbf{P}) + \mathbf{P}.
\end{equation}
The network consists of a two-layer multi-layer perceptron (MLP) with a residual connection.
In training, both $\mathbf{P}$ and $\phi_m$ are optimized jointly. Once the training is completed, $\mathbf{P'}$ can be computed according to Eq. \eqref{eq6} and cached for inference. 
There is no need to maintain $\mathbf{P}$ and $\phi_m$ for inference. 
Hence, the reparameterization introduces no extra computation and storage costs for inference.

\section{Experiments}
\label{sec:exp}
\vspace{-3mm}
\subsection{Dataset and Evaluation Metric}
\begin{table}[t]
	\caption{\textit{WER (\%) evaluation w.r.t. varying prompt length and reparameterization methods. DP denotes deep prompting, sharedMLP denotes we use one shared MLP to parameterize all the prompts, and sepMLP denotes we use separate MLPs for each layer. All the results reported here are evaluate on Libri2Mix \texttt{dev-clean}.}} 
	\centering
	\label{tab:1}
    \setlength{\tabcolsep}{1.6mm}{
	\begin{tabular}{lcccccc}
		\toprule
		\multirow{2}*{\textbf{Method}}  & \multicolumn{6}{c}{\textbf{Prompt length}}  \\
		\cmidrule(lr){2-7}
		  & 4 & 8 & 16 & 32 & 64 & avg.\\
		\midrule
		DP  & 13.90 & 13.97 & 12.54 & 14.55 & 12.52 & 13.50  \\
		DP + sharedMLP & 14.30 & 14.17 & 14.18 & 13.27 & 14.82  & 14.15 \\
        DP + sepMLP & 14.59 & 12.18 & 12.37 & 12.68 & 12.17& {\bf 12.80}   \\
		\bottomrule
	\end{tabular}}
 \vspace{-7mm}
\end{table}

We evaluate our proposed methods on Libri2Mix \cite{cosentino2020librimix}, which consists of overlapped speech mixed by pairs of different speakers' utterances from LibriSpeech \cite{panayotov2015librispeech} at a signal-to-noise ratio (SNR) (in dB) randomly sampled from $\mathcal{N}(0, 4.1^2)$. All speech mixtures were sampled at 16kHz and constructed in \textit{max} mode, where the shorter utterance is zero-padded to match the length of the longer one and aligned to the starting sample, forming the highly overlapped speech.

In our experiments, soft prompts are optimized on \texttt{train-100} \texttt{-clean} and \texttt{train-100-both} where WHAM! \cite{wichern2019wham} is included as background noise. Evaluations are conducted on the clean subsets \texttt{dev-} and \texttt{test-clean} and the noisy subsets \texttt{dev-} and \texttt{test-both}. Both training sets contain 13,900 utterances for a total of 58 hours. Each of the evaluation sets comprises 3,000 utterances, amounting to a total of 11 hours. 

To obtain the target speaker embeddings, we follow the prior work \cite{tse4_baseline} to use a 15s utterance for each speaker from the Libri-Light \cite{kahn2020libri} project, which is not overlapped with LibriSpeech, to extract the 512-dimension target speaker x-vectors \cite{xvec}.

We use the word error rate (WER) as the performance metric. 
Both the model prediction and ground truth texts are normalized by following the way in \cite{whisper}
before computing WER.

\subsection{Training Configuration}
We evaluate the performance of our prompting scheme on three models: Whisper-large v2, Whisper-medium, and Whisper-small.
In the training phase, we fix all the parameters of Whisper and optimize the soft prompts, the speaker projection layer, and the reparameterization MLP (if it exists) jointly using the AdamW optimizer \cite{adamW} with an initial learning rate of 1e-4. 
We train a total of 10 epochs on \texttt{train-100-clean} and another epoch on \texttt{train-100-both}.
After the first 5 epochs, we reduce the learning rate to 1e-5. The training process is performed on a small server with one single NVIDIA RTX3090 24GB. For all experiments, We keep random seeds the same for fair comparison.
\vspace{-3mm}
\subsection{Main Results}

\subsubsection{Selection of Prompt Length and Reparameterization Method}
\begin{table}[!t]
    \centering
    \caption{\textit{Evaluation of proposed prompting methods and comparison with state-of-the-art full fine-tuning methods on Libri2Mix \texttt{test-clean} and \texttt{test-both}. \dag: We do not report results with JSM module in \cite{tse4_baseline} for the assumption that the interference speaker identity is unknown in a speech mixture.}}
    \setlength{\tabcolsep}{0.7mm}{
    \begin{tabular}{lcccc}
    \toprule
    \multirow{2}*{\textbf{Method}}& \multicolumn{2}{c}{\textbf{WER (\%)}}&\multicolumn{2}{c}{\textbf{Task params}}\\
    \cmidrule(lr){2-3} \cmidrule(lr){4-5}
    & \texttt{clean} & \texttt{both} & \textit{Train} & \textit{Infer} \\
    \midrule
    WavLM Base$^{+}$ + TSE\cite{tse4_baseline}$^{\dag}$&{12.32}&-&95.34M&95.34M \\
    WavLM Base + cLN\cite{tse5_baseline}&-&27.5&95.34M&95.34M\\
    TS-HuBERT + cLN\cite{tse5_baseline}&-&{\bf 24.8}&105.18M&105.18M\\
    \midrule
    Whisper-small + DP&24.62&45.46&{\bf 0.69M}&{\bf 0.69M}\\
    \hspace{1.8cm} ++ MLP&23.08&44.16&14.91M&{\bf 0.69M}\\
    Whisper-small + LoRA &19.06&36.82&0.89M&0.89M\\
    Whisper-medium + DP&13.89&29.81&{\bf 1.31M}&{\bf 1.31M}\\
    \hspace{1.8cm} ++ MLP&13.54&30.72&51.82M&{\bf 1.31M}\\
    Whisper-medium + LoRA &{\bf 11.98}&26.38&1.85M&1.85M\\
    Whisper-large + DP&14.82&30.19&{\bf 1.97M}&{\bf 1.97M}\\
    \hspace{1.8cm} ++ MLP&14.78&30.71&107.11M&{\bf 1.97M}\\
    Whisper-large + LoRA &13.07&26.51&2.87M&2.87M\\
    \midrule
    Whisper-small + fine-tuning&23.18&41.39&241.37M&241.37M \\
    Whisper-medium + fine-tuning&18.79&34.04&762.85M&762.85M \\
    \bottomrule
    \end{tabular}}
    \label{tab:my_label}
    \vspace{-7mm}
\end{table}

We first analyze how the prompt length and reparameterization method affect the performance. We choose \textit{Whisper-large v2} as the foundation model and extend it to TS-ASR using deep prompting while varying prompt lengths within \{4, 8, 16, 32, 64\}.
Additionally, we evaluate the performance of two reparameterization schemes: (1) \textit{sharedMLP} that employs a shared MLP to reparameterize all prompts; and (2) \textit{sepMLP} that uniquely employs separate MLPs for each layer. As a point of reference for benchmarking, we also examine a scheme that does not involve the utilization of reparameterization MLPs.

The results are shown in Table~\ref{tab:1}. 
As the prompt length increases, WER tends to decrease. When the length exceeds 16, the performance gain becomes less prominent. 
To strike a balance between performance and parameter efficiency, we opt to set the prompt length to 16 for all subsequent experiments.
Table \ref{tab:1} also shows how the three reparameterization schemes influence the performance.
A shared reparameterization MLP is even worse than the scheme without reparameterization. This could be attributed to that the shared MLP restricts the learning of inter-layer variations. 
This is proved by the fact that adopting separate MLPs for each layer leads to performance enhancements. As a result, we opt to employ separate MLPs for the reparameterization of soft prompts in our subsequent experiments. 
\vspace{-3mm}
\subsubsection{SOTA Comparison}

We compare our prompting scheme with other SOTA methods in two aspects: performance (evaluated in terms of WER) and parameter efficiency (evaluated in terms of task parameters). 
The results are shown in Table \ref{tab:my_label}. 
While the prior works based on WavLM and TS-HuBERT have slightly better performance, the training is expensive and requires extensive structure modifications. Deep prompting, in contrast, is a much cheaper choice to extend Whisper for TS-ASR. We also report the results of LoRA, which is another PEFT method for adapting large models. It achieves the best performance, as shown in Table \ref{tab:my_label}. However, despite performance, prompting methods have their unique advantages (e.g., supports multi-task parallel inference \cite{prompt}). 
In deep prompting, employing reparameterization MLP gives an additional performance improvement on \texttt{test-clean} at the expense of an increase in the number of parameters to tune. 
However, once the training is completed, the MLP can be simply discarded. 
In this way, the number of task-specific parameters to be stored for inference remains limited.

We then report the results of full fine-tuning in the last two rows of Table \ref{tab:my_label}. Due to computing resource constraints, we are unable to provide the results for Whisper-large. A noticeable trend emerges when comparing full fine-tuning with the prompting approach: despite a substantial increase in the number of tunable parameters, full fine-tuning fails to exhibit satisfactory performance. The underlying cause of this performance gap is attributed to the training set \texttt{train-100-clean} is relatively small, causing severe over-fitting during full fine-tuning. 
This stark contrast underscores the effectiveness of prompt tuning in mitigating overfitting risks, especially when adaptation data for large-scale models are limited.
\vspace{-3mm}
\subsubsection{Ablation Study}
\begin{table}[t] 
    \centering
	\caption{\textit{Ablation study of our prompting scheme. Enc./Dec. denote inserting prompts in encoder/decoder blocks, PT denotes prompt tuning, DP denotes deep prompting, MLP denotes reparameterization feed-forward network and L/M/S denotes Whisper-large/medium/small. All results are reported in terms of WER (\%).}} 
    	\label{tab:2}
     \setlength{\tabcolsep}{0.7mm}{
	\begin{tabular}{ccccccccccc}
		\toprule
		\multirow{2}*{\textbf{Enc.}}& \multirow{2}*{\textbf{Dec.}} & \multirow{2}*{\textbf{PT}} & \multirow{2}*{\textbf{DP}} & \multirow{2}*{\textbf{MLP}} &\multicolumn{3}{c}{\texttt{test-clean}} & \multicolumn{3}{c}{\texttt{test-both}}\\
            \cmidrule(lr){6-8} \cmidrule(lr){9-11}
        &&&&&\textbf{L}&\textbf{M}&\textbf{S}&\textbf{L}&\textbf{M}&\textbf{S} \\
		\midrule
		\ding{55}&\ding{55}&\ding{55}&\ding{55}&\ding{55}&78.92&85.42&92.24&83.19&86.89&99.43\\
		\checkmark&\checkmark&\checkmark&\ding{55}&\ding{55}&17.59&17.73&31.03&37.26&39.38&56.30 \\
		\checkmark&\checkmark&\checkmark&\ding{55}&\checkmark&15.92&14.61&24.27&32.25&32.34&45.72\\
		\checkmark&\checkmark&\checkmark&\checkmark&\ding{55}&14.82&13.89&24.62&{\bf 30.19}&{\bf 29.81}&45.46\\
        \checkmark&\checkmark&\checkmark&\checkmark&\checkmark&{\bf 14.78}&{\bf 13.54}&{\bf 23.08}&30.71&30.72&{\bf 44.16} \\
        \midrule
        \checkmark&\ding{55}&\checkmark&\checkmark&\checkmark&17.24&15.43&24.28&33.30&31.95&44.30 \\
        \ding{55}&\checkmark&\checkmark&\checkmark&\checkmark&35.09&43.96&58.83&52.03&62.26&80.14 \\
		\bottomrule
	\end{tabular}}
 \vspace{-7mm}
\end{table}
We conduct ablation experiments to verify the impact of design components, including prompt tuning, deep prompting, and reparameterization MLP. As shown in Table \ref{tab:2}, Whisper without any modification (i.e. zero-shot configuration) cannot perform well on target-speaker ASR. 
When we insert target speaker embedding together with soft prompts into both encoder and decoder input, a dramatic WER reduction, e.g. from 78.92\% to 17.59\% for Whisper-large on \texttt{test-clean}, is observed.
This shows the efficacy of prompt tuning. 
By further adding the reparameterization MLP module, we observe performance improvement in most of the cases, showing that reparameterization of soft prompts can benefit the training. 
In addition, configurations with deep prompting show consistent performance improvement.
This shows that more tunable parameters bring better performance and the soft prompts closer to the output layer have a more critical performance effect. 

We further investigate how prompting location affects performance. Specifically, based on the best configuration (i.e., Whisper + PT + DP + MLP),  we ablate the soft prompts inserted in the encoder/decoder to test the influence of varying prompting locations.
Hereby, we set the prompt length to 32 to keep the number of parameters constant. The results are shown in the last two rows of Table \ref{tab:2}, from which we see that the soft prompts inserted in the encoder play an important role in guiding the model while those inserted in the decoder play a relatively minor role but still lead to performance improvement. So, inserting soft prompts both in the encoder and decoder will lead to the best performance with a constant number of parameters.
\vspace{-3mm}
\subsubsection{Retaining Whisper's Featured Abilities}
We anticipate a practical TS-ASR system to not only accurately transcribe the desired speech of a target speaker but also to execute the entire speech processing pipeline. This includes various post-processing tasks, such as inverse text normalization and timestamp tagging, among others. The original Whisper has already demonstrated commendable performance in these aspects. However, it's important to note that the prompted Whisper can no longer showcase its automatic punctuating and capitalizing abilities while performing TS-ASR. This limitation arises because the ground truth texts in LibriSpeech, which are \textit{manually labeled}, lack case sensitivity and punctuation. Soft prompts have learned and perpetuated this bias. To retain the automatic inverse text normalization ability of Whisper, we propose to employ the \textit{auto-labeled} texts generated by the original Whisper model on single-talker utterances as ground truth to supervise the training of soft prompts.
\begin{table}[h]
    \fontsize{6pt}{1pt}
    \it
    \centering
    \setlength{\tabcolsep}{1mm}{
    \vspace{-2mm}
    \begin{tabular}{|l|}
    \hline
        bartley started when hilda rang the little bell beside her dear me why did you do that\\
        \hline
        \textcolor{green}{B}artley started when \textcolor{green}{H}ilda rang the little bell beside her\textcolor{red}{.} \textcolor{green}{D}ear me\textcolor{red}{,} why did you do that\textcolor{red}{?} \\
        \hline
        \textcolor{blue}{$\langle|0.00|\rangle$}Bartley ... beside her.  \textcolor{blue}{$\langle|3.26|\rangle$ $\langle|3.26|\rangle$}Dear me ... do that?  \textcolor{blue}{$\langle|5.02|\rangle$}\\

    \hline
    \end{tabular}}
    \vspace{-3mm}
    \label{tab4}
\end{table}

The above table shows the results with and without inverse text normalization, where we highlight the automatically inserted punctuation and capitalized letters. We see that through training soft prompts supervised by the auto-labeled texts that are inverse-text-normalized, the prompted Whisper can now automatically perform inverse-text normalization along with target-speaker ASR with virtually no performance loss. By further removing the special token \textit{$\langle|$no-timestamps$|\rangle$} in multitask format (see Sec. \ref{sec2.1}), the prompted Whisper can produce reliable timestamps, as indicated by the highlights in the last row of the table, even though timestamps were not included in the adaptation data.

\section{Conclusion}
\label{sec:typestyle}
In this work, we have proposed a prompting scheme, that jointly leverages prompt tuning, and deep prompting together with the reparameterization technique, to extend the single-talker ASR model to the target-speaker ASR of multi-talker overlapped utterances. 
Our proposed method has shown comparable performance with SOTA and such performance was achieved with an attractively small amount of task parameters compared to existing methods. 
Notably, our method also retained the automatic inverse text normalization and timestamp prediction capabilities of the original Whisper model. This ensures that the generated transcriptions remain both natural and informative, further enhancing the versatility and utility of our method.



\bibliographystyle{IEEEbib}
\bibliography{refs}

\end{document}